\documentclass{article}
\pdfoutput=1
\usepackage{spconf,amsmath,graphicx}

\usepackage{times}
\usepackage{epsfig}
\usepackage{graphicx}
\usepackage{amsmath}
\usepackage{amssymb}
\usepackage{booktabs}
\usepackage{enumitem}
\usepackage{soul}
\usepackage{mathtools}

\usepackage{color}
\usepackage[para]{footmisc}
\usepackage{multirow}
\usepackage{diagbox}

\newcommand{\keypoint}[1]{\vspace{0.1cm}\noindent\textbf{#1}\quad}
\newcommand{\cut}[1]{}
\newcommand{\etal}{\textit{et. al. }}
\usepackage{algorithm}
\usepackage{algpseudocode}

\usepackage[nopar]{lipsum}
\usepackage{arydshln}
\usepackage{color, colortbl}
\definecolor{LightCyan}{rgb}{0.95,0.95,0.95}
\definecolor{Black}{rgb}{0,0,0}
\newcolumntype{a}{>{\columncolor{LightCyan}}c}
\usepackage[breaklinks,colorlinks]{hyperref}
\usepackage{adjustbox}
\usepackage{color}
\usepackage{tabularray}

\title{EXPLORING SELF-SUPERVISED REPRESENTATION LEARNING FOR LOW-RESOURCE MEDICAL IMAGE ANALYSIS}
\name{Soumitri Chattopadhyay$^{1}$, Soham Ganguly$^{*1}$, Sreejit Chaudhury$^{*1}$, Sayan Nag$^{*2}$, Samiran Chattopadhyay$^{1}$}
\address{$^{1}$Jadavpur University \qquad $^{2}$University of Toronto}

\begin{document}
%
\maketitle

\begin{abstract}
The success of self-supervised learning (SSL) has mostly been attributed to the availability of unlabeled yet large-scale datasets. However, in a specialized domain such as medical imaging which is a lot different from natural images, the assumption of data availability is unrealistic and impractical, as the data itself is scanty and found in small databases, collected for specific prognosis tasks. To this end, we seek to investigate the applicability of self-supervised learning algorithms on small-scale medical imaging datasets. In particular, we evaluate $4$ state-of-the-art SSL methods on three publicly accessible \emph{small} medical imaging datasets. Our investigation reveals that in-domain low-resource SSL pre-training can yield competitive performance to transfer learning from large-scale datasets (such as ImageNet). Furthermore, we extensively analyse our empirical findings to provide valuable insights that can motivate for further research towards circumventing the need for pre-training on a large image corpus. To the best of our knowledge, this is the first attempt to holistically explore self-supervision on low-resource medical datasets. Source codes are available at: \href{https://github.com/soumitri2001/SmallDataSSL}{https://github.com/soumitri2001/SmallDataSSL}  

%

\end{abstract}

\def\thefootnote{*}\footnotetext{\hspace{-1.5mm}denotes equal contribution.}\def\thefootnote{\arabic{footnote}}
\begin{keywords}
Self-supervised learning, low-resource, medical image analysis, contrastive learning, non-contrastive
\end{keywords}
\section{Introduction}
\label{sec:intro}

With its label efficiency and the ability to learn robust representations, self-supervised learning (SSL) \cite{chen2020simple, chen2021exploring, bardes2021vicreg, yeh2022decoupled, manna2022swis} has shown promising performance across various visual understanding tasks. The typical paradigm for SSL has been to perform pre-training on a large-scale dataset (such as ImageNet \cite{deng2009imagenet}), followed by application to the downstream task with in-domain data. This, however, poses a fundamental challenge -- there is often a dearth of available data for pre-training in a domain that is very different from commonly available ones. One such field is medical image analysis, which comprises disease detection \cite{manna2022self}, classification \cite{azizi2021big} and segmentation \cite{basak2022addressing} from medical scan images. Understandably, medical images vary greatly, not only across anatomical regions, but also in their capturing mechanisms. Moreover, due to privacy issues (and otherwise), medical data is hardly available on a large scale; they are mostly collected in small-scale databases private to a geographical location or laboratory. This is more prominent for specialized prognosis cases (such as histopathological scans for cancer detection). The need for in-domain training thereby arises in such cases. One might argue that supervised in-domain learning is a possible solution; however, it must be noted that deep learning models are highly prone to overfitting when it comes to small-scale datasets. SSL techniques, on the other hand, can learn very robust features without the need for labels, making them immune to such problems \cite{chen2020simple, chen2021intriguing}. Furthermore, medical image annotation is a cumbersome process requiring a lot of effort from skilled clinicians, further motivating the use of self-supervision for medical imaging tasks \cite{shurrab2022self, cheplygina2019not, chen2019self}.    

While SSL has been leveraged to large-scale medical datasets \cite{sowrirajan2021moco, azizi2021big, ghesu2022self}, there has been no prior work that specifically focuses on self-supervision for smaller medical datasets. In fact, even in the broader vision community, only a couple of works \cite{wallace2020extending, cao2021rethinking} to date have approached self-supervision in low-resource settings. The work by Wallace \etal \cite{wallace2020extending}, which evaluated some of the pretext-task-based SSL algorithms \cite{pathak2016context, noroozi2016unsupervised, wu2018unsupervised} over a wide range of visual domains, was one of the earliest. However, with recent (non-)contrastive approaches achieving state-of-the-art performance on vision tasks, mere evaluation of classical pretext tasks alone is rather outdated. 

To this end, we explore the applicability of SSL algorithms on small medical imaging datasets. Specifically, our study encompasses both contrastive (CL) and non-contrastive (NCL) learning algorithms in literature, and uses datasets \cite{spanhol2016breast, kather2016CRCH, kermany2018labeled} that have been proposed for clinical purposes and not for mere benchmarking. We further analyse the findings with suitable ablations to present insights and trends across datasets, which can open up new research directions. Last but not the least, the empirical results of our paper can be directly referred to as a benchmark by other researchers working on similar lines. Our study is motivated by the recent work of Cao \etal \cite{cao2021rethinking}, which approached SSL in a scaled-down fashion from both model and data aspects. However, it should be noted that we deal entirely with medical image data and our study includes recent non-contrastive SSL variants as well, which have been overlooked in \cite{cao2021rethinking}.



Summing up, the salient contributions of our work are: 
\begin{enumerate}
    \item We extensively evaluate $4$ recent state-of-the-art SSL algorithms on three real-world small-scale medical image classification datasets.
    \item We analyse our findings across various ablations and experimental setups to present several insights and noteworthy trends, which ought to pave the way for future studies.
    \item To the best of our knowledge, this is one of the first attempts to explore both data and label efficiency together in the context of medical image analysis.
\end{enumerate}

\begin{figure}[tbp]
    \centering
    \includegraphics[width=\linewidth]{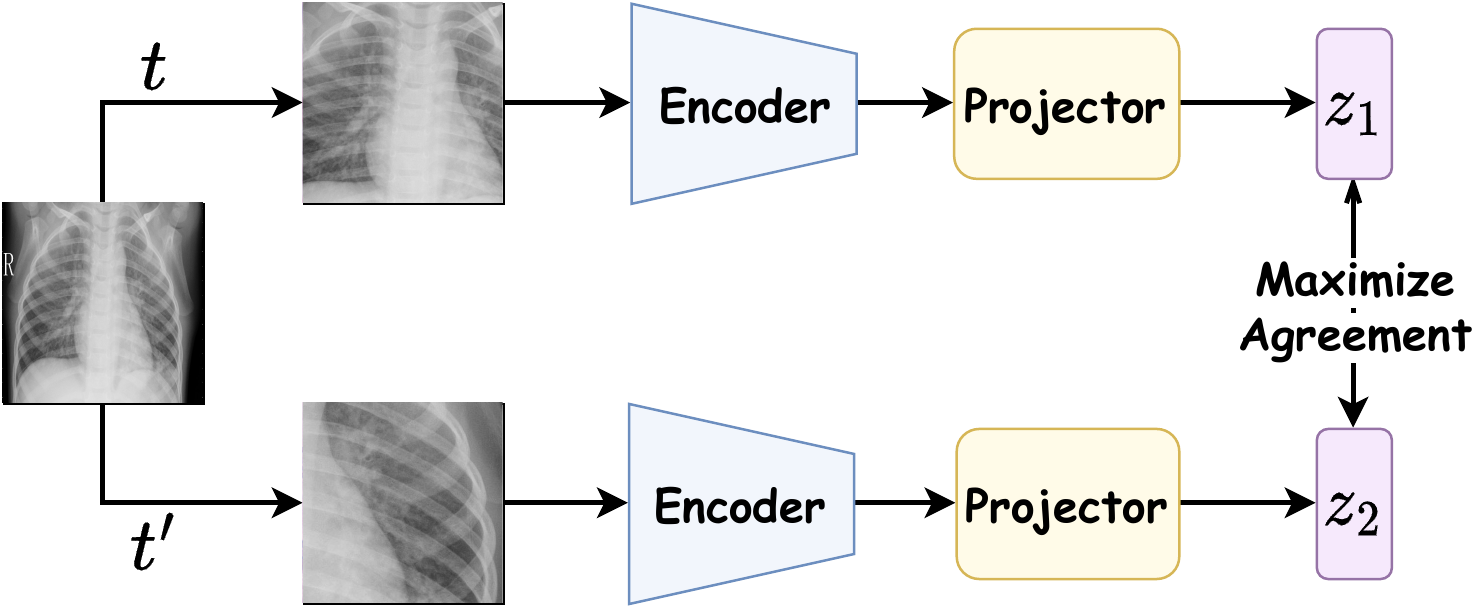}
    \caption{Generic representation of self-supervised learning.}
    \label{fig:mainfig}
\end{figure}

\section{Materials and Methods}
\label{sec:methods}

\subsection{Self-Supervised Learning Algorithms}
In this paper, we consider two families of SSL algorithms, namely contrastive \cite{chen2020simple, jaiswal2020survey} and non-contrastive \cite{bardes2021vicreg, chen2021exploring} learning, which have yielded state-of-the-art performance at various vision tasks, surpassing traditional SSL works such as inpainting \cite{pathak2016context} or jigsaw puzzle \cite{noroozi2016unsupervised}.

\keypoint{Contrastive Learning:}These algorithms \cite{chen2020simple, he2020momentum, yeh2022decoupled} aim at pulling the representations of similar samples i.e. positive pairs closer while simultaneously pushing away dissimilar samples (i.e. negative pairs). In its typical setup \cite{chen2020simple, he2020momentum}, two augmented views of the same image $x_i$ are passed through a shared CNN encoder and its cascaded projector head, with the representations obtained being $z_i^{(1)}$ and $z_i^{(2)}$. For each augmented representation $z_i^{(m)}$, the contrastive loss objective sets the other corresponding augmentation $z_i^{n}$ as the positive label $(m{\neq}n)$. Having thus reduced to a classification problem, the objective is formulated as a cross-entropy loss function with respect to $z_i^{(m)}$, also known as the NT-Xent loss \cite{sohn2016improved}. The overall loss would be a sum over all the individual losses, given by Eq. \ref{eq:cl1}, with temperature parameter $\tau$.

\begin{equation}\label{eq:cl1}
\resizebox{\columnwidth}{!}{
    $\mathbf{L}_{CL} = -\sum\limits_{\substack{m\in\{1,2\} \\ i\in\{1...B\}}}\log \frac{\exp(z_i^{(1)}\cdot z_i^{(2)}/\tau)}{\exp(z_i^{(1)}\cdot z_i^{(2)}/\tau) + \sum_{j \neq i}\exp(z_i^{(m)}\cdot z_j^{(p)}/\tau)}$
    }
\end{equation}

In this work, we have chosen SimCLR \cite{chen2020simple} and DCL \cite{yeh2022decoupled} as the representative contrastive SSL methods, which have shown state-of-the-art performance across various high-resource vision tasks. While SimCLR is a straightforward implementation of the NT-Xent contrastive loss (Eq. \ref{eq:cl1}), DCL removes the negative-positive coupling multiplier term \cite{yeh2022decoupled} and formulates a decoupled version of the loss.


\keypoint{Non-Contrastive Learning:}Contrastive SSL methods require large mini-batch sizes for increased negative pairs (SimCLR \cite{chen2020simple}) and/or large memory banks (MoCo \cite{he2020momentum}). Such a scaling in batch size and number of overall samples leads to smaller generalization gaps which boost performance \cite{pham2021combined}. However, such requirements make Contrastive SSL resource-inefficient. To alleviate this, NCL methods such as SimSiam \cite{chen2021exploring} and VICReg \cite{bardes2021vicreg} have been introduced in recent literature. These methods do not require any \textit{negative pairs} and therefore perform well even with smaller mini-batch sizes. In particular, they attempt to minimize redundancy (information collapse) between components of the encoded representations generated from two views of the input. VICReg does so by an information maximization paradigm involving a variance-preservation term \cite{bardes2021vicreg} for the embedded components. On the other hand, SimSiam \cite{chen2021exploring} minimizes the negative cosine similarity of one view with the projector output of the other. It calculates a symmetric loss while applying a stop gradient to the non-projected view as follows:

\begin{gather}
\mathbf{L}_{SimSiam} = \frac{1}{2}{sim}(p_{1},stgrad(z_{2})) + 
\frac{1}{2}{sim}(p_{2},stgrad(z_{1}))
\end{gather}
 Here, $sim(\cdot)$ is the negative cosine similarity function given by Eq, \ref{eq:sim}, while $stgrad(\cdot)$ refers to gradient stopping operation.
 \begin{equation}\label{eq:sim}
     sim(u,v) = -\frac{u}{\|u\|_2}\cdot\frac{v}{\|v\|_2}
 \end{equation}

In our study, we have adopted VICReg and SimSiam as the representative NCL algorithms.


\subsection{Datasets}

Three publicly accessible small-scale medical imaging datasets \cite{spanhol2016breast, kather2016CRCH, kermany2018labeled} have been used in this study. Note that all of these datasets were proposed keeping clinical use in mind, which ensures the practicality of our investigation.

\keypoint{Breast:}The BreaKHis400x \cite{spanhol2016breast} is a breast histopathological dataset comprising microscopic biopsy images of 588 benign and 1232 malignant breast tumors at 400x magnification.

\keypoint{Colorectal:}This is a class-balanced histopathological dataset comprising 5000 tissue images evenly divided among 8 classes \cite{kather2016CRCH}, proposed for the detection of colorectal cancer.

\keypoint{Pneumonia:}This is a lung X-ray dataset by Kermany \etal \cite{kermany2018labeled} comprising 5856 images, out of which 1583 are ``Normal'' while 4273 belong to class ``Pneumonia''.

\begin{figure}
    \centering
    \includegraphics[width=1\linewidth]{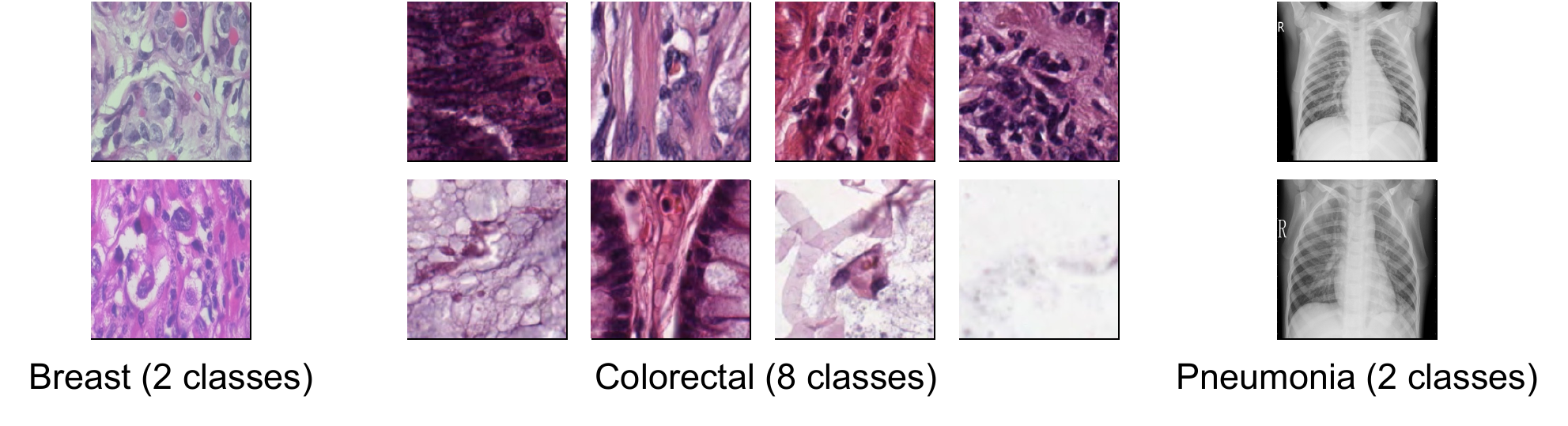}
    \caption{Sample images from each of the datasets used.}
    \label{fig:samples}
\end{figure}

\noindent
Each of the datasets was randomly split into train, validation and test sets in the ratios 70\%, 10\% and 20\% respectively. Some representative sample images are shown in \autoref{fig:samples}.

\section{EXPERIMENTAL SETUP}\label{sec:setup}
\subsection{Implementation details}
\keypoint{Pre-training:}We used a randomly initialized ResNet50 encoder, a standard architecture used for SSL training \cite{chen2020simple, yeh2022decoupled}. For SimCLR and DCL, (i.e. contrastive), a $2$-layer MLP projector head with $128$-dimensional output \cite{chen2020simple} was cascaded with the encoder. For SimSiam \cite{chen2021exploring}and VICReg \cite{bardes2021vicreg}, we followed their original configurations -- $2$-layer MLP projector/expander heads with outputs of $512$ and $4096$ dimensions respectively. Batch sizes of $16$, $32$ and $64$ were considered, learning rate of $0.01$, SGD optimizer and training for $250$ epochs.

\keypoint{Downstream:}We used linear evaluation, wherein we froze the pre-trained encoder and trained a linear classifier on top of the frozen representations. Following the literature \cite{chen2021exploring, yeh2022decoupled}, we attached a $2$-layered MLP classifier and trained it for $100$ epochs using Adam optimizer with a StepLR scheduler $(\gamma = 0.5)$ decaying over episodes of $25$ epochs. The best individual hyperparameter settings obtained from careful tuning have been provided in \autoref{tab:hp}.

\begin{table}[!h]
\centering
\resizebox{1\columnwidth}{!}{\begin{tabular}{
  |c|c|c|c|
}
\hline
\diagbox{\textbf{Method}}{\textbf{Dataset}} & \textbf{Breast}      & \textbf{Colorectal}         & \textbf{Pneumonia}   \\ \hline 
\textbf{DCL}              & 16(0.01) & 16(0.01) & 64(0.5) \\ \hline 
\textbf{SimCLR}            & 16(0.01) & 16(0.005) & 64(0.1) \\ \hline 
\textbf{SimSiam}           & 16(0.01) & 16(0.01) & 16(0.01) \\ \hline 
\textbf{VICReg}            & 16(0.005) & 16(0.005) & 16(0.01) \\ \hline  
\end{tabular}}
\caption{Optimal hyperparameters for linear evaluation. The values have been reported as $Batch Size (Learning Rate)$.}
\label{tab:hp}
\end{table}

All of the models were implemented in PyTorch \cite{paszke2019pytorch} and the experiments were conducted on a 16GB Nvidia Tesla V100 GPU. We shall release our codes upon acceptance.

\subsection{Evaluation}
\keypoint{Protocol:}We used two types of evaluation -- (1) using a KNN classifier directly on the frozen pre-trained representations and (2) linear probing. The purpose of KNN evaluation is to evaluate the robustness of the SSL features without any further training/fine-tuning, whereas linear probing follows supervised training of an MLP classifier on top of the frozen backbone. For KNN, the value of $k$ was set to 200, following recent literature \cite{yeh2022decoupled}.

\keypoint{Metrics:}$Accuracy$ and weighted $F1$-$score$, the two most commonly used classification metrics, have been used in this study to evaluate the performance of SSL algorithms on the mentioned small-scale medical imaging datasets.

\section{Results and Analysis}\label{sec:analysis}

We report the performances of the SSL algorithms on the three medical image datasets using the metrics described above. The best results of KNN-based evaluation are found at a batch size of $64$, which have been tabulated in \autoref{tab:knn}.

\begin{table}[!h]

\resizebox{1\columnwidth}{!}{\begin{tabular}{|cl c c c|}
\toprule

\multicolumn{2}{|c|}{\multirow{2}{*}{
\begin{tabular}[c]{@{}c@{}}\diagbox{\textbf{Method}}{\textbf{Dataset}} \\ \end{tabular}}} & \multicolumn{1}{c|}{\multirow{2}{*}{\textbf{Breast}}}              & \multicolumn{1}{c|}{\multirow{2}{*}{\textbf{Colorectal}}}                 & \multicolumn{1}{c|}{\multirow{2}{*}{\textbf{Pneumonia}}}           \\ 
\multicolumn{2}{|c|}{}                                                                             &     &   & \\
\midrule
\multicolumn{2}{|c|}{\textbf{DCL} \cite{yeh2022decoupled}}                                                                         & 0.837 / 0.829 & \textbf{0.796 / 0.795} &    0.771 / 0.759 \\ 
\multicolumn{2}{|c|}{\textbf{SimCLR} \cite{chen2020simple}}                                                                       & 0.835 / 0.828 & 0.792 / 0.790            & 0.766 / 0.748 \\ \midrule
\multicolumn{2}{|c|}{\textbf{SimSiam} \cite{chen2021exploring}}                                                                      & \textbf{0.842 / 0.836} & 0.746 / 0.739 &              0.715 / 0.676 \\ 
\multicolumn{2}{|c|}{\textbf{VICReg} \cite{bardes2021vicreg}}                                                                      & 0.838 / 0.831 & 0.784 / 0.780            & \textbf{0.787 / 0.773} \\ \bottomrule
\end{tabular}}

\caption{KNN evaluation results. Scores have been reported as $accuracy / F1$-$score$.}
\label{tab:knn}

\end{table}


As for the linear evaluation setup, the best hyperparameter settings have been reported in \autoref{tab:hp}, based on which the linear evaluation results are tabulated in \autoref{tab:lin}. For reference, we also experimented with a linear-like transfer learning setup where the frozen backbone comprises supervised ImageNet \cite{deng2009imagenet} trained representations (denoted by ImgNet+TL). 


\begin{table}[!h]
\centering
\resizebox{1\columnwidth}{!}{\begin{tabular}{
  |c|c|c|c|
}
\toprule
\diagbox{\textbf{Method}}{\textbf{Dataset}} & \textbf{Breast}      & \textbf{Colorectal}         & \textbf{Pneumonia}   \\ \midrule 
\textbf{DCL} \cite{yeh2022decoupled}              & 0.939 / 0.939 & \textbf{0.921 / 0.921} & 0.790 / 0.773 \\
\textbf{SimCLR} \cite{chen2020simple}           & 0.925 / 0.924 & 0.874 / 0.874 & 0.761 / 0.734 \\ \midrule 
\textbf{SimSiam} \cite{chen2021exploring}          & 0.908 / 0.907 & 0.898 / 0.898 & 0.776 / 0.754 \\ 
\textbf{VICReg} \cite{bardes2021vicreg}           & \textbf{0.939 / 0.939} & 0.915 / 0.915 & \textbf{0.817 / 0.806} \\ 
\midrule  
\textbf{ImgNet+TL}        & 0.916 / 0.915 & 0.911 / 0.910 & 0.798 / 0.775 \\
\bottomrule
\end{tabular}}
\caption{Linear evaluation results. Scores have been reported as $accuracy / F1$-$score$.}
\label{tab:lin}
\end{table}

\begin{figure*}[t]
    \centering
    \includegraphics[width=\linewidth]{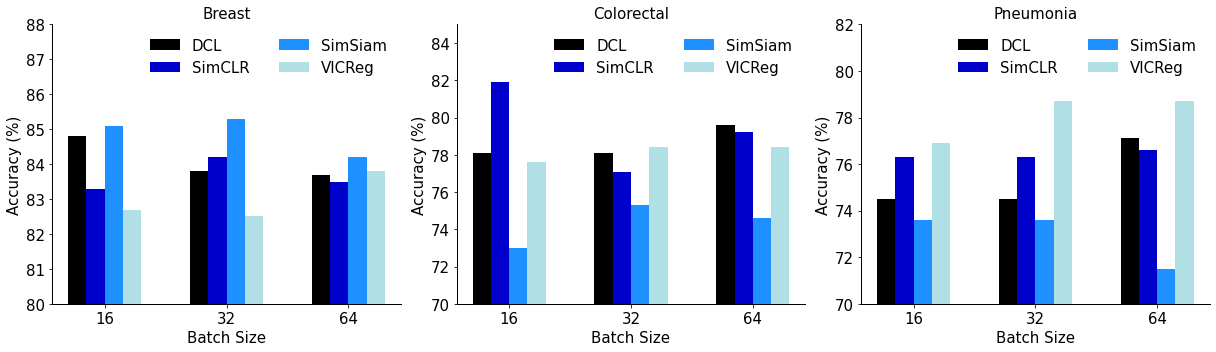}
    \caption{KNN evaluation $Accuracy$ (\%) vs Batch Size for 4 different methods across all 3 datasets.}
    \label{fig:ablation}
\end{figure*}

From the empirical results obtained, we now further analyze their implications along various axes of comparison.

\keypoint{Contrastive vs Non-contrastive:}
An important point of investigation would be how the two SSL paradigms  evaluated in this paper, namely CL and NCL, perform at medical image classification under low data setting. From the KNN evaluations in \autoref{tab:knn}, we observe that the NCL variants consistently perform at par with the CL approaches, outperforming the latter in Breast and Pneumonia datasets. The trend is repeated in the linear probing results (\autoref{tab:lin}), where VICReg outperforms both DCL and SimCLR on the above mentioned datasets, while being very competitive to DCL on the Colorectal dataset. We attribute this to the fact that NCL learning algorithms can work effectively with fewer number of data samples, since there is no explicit need for a higher number of samples, as in the case of CL methods \cite{chen2020simple,chen2021intriguing}. Thus, given a fixed batch-size that is \textit{not} sufficiently large, NCL methods would yield superior results compared to CL ones.

\keypoint{Effect of batch size:}It is well-known that SSL approaches are influenced by the batch size taken. For CL, higher batch size ensures increase in negative samples \cite{chen2021intriguing, yeh2022decoupled}, while in NCL methods, feature normalization is done along the batch dimension \cite{bardes2021vicreg}. To investigate these trends for small-scale medical data, we ablated the batch size across $16$, $32$ and $64$ and plotted the KNN accuracy scores in \autoref{fig:ablation}. Note that batch size of $128$ was infeasible due to resource constraints. 

CL methods are expected to show improvement with increase in batch size \cite{chen2020simple, chen2021intriguing} due to the rise in negative samples. Surprisingly, we do not see such a generalised trend for either of SimCLR or DCL. However, from \autoref{fig:ablation}, it is evident that VICReg shows improvement with increase in batch sizes, which is in accordance with the findings of the original paper \cite{bardes2021vicreg}. This can be attributed to the feature normalization operation in its loss function, where a larger batch size would ensure less noisy gradients, thereby improving generalisability. On the other hand, for SimSiam, the peak is usually at an intermediate value, asserting its effectiveness under lower batch size configurations \cite{chen2021exploring}. This conveys that non-contrastive SSL approaches are better suited for low-resource medical imaging compared to CL methods.

\begin{figure}[!h]
    \centering
    \includegraphics[width=1\columnwidth]{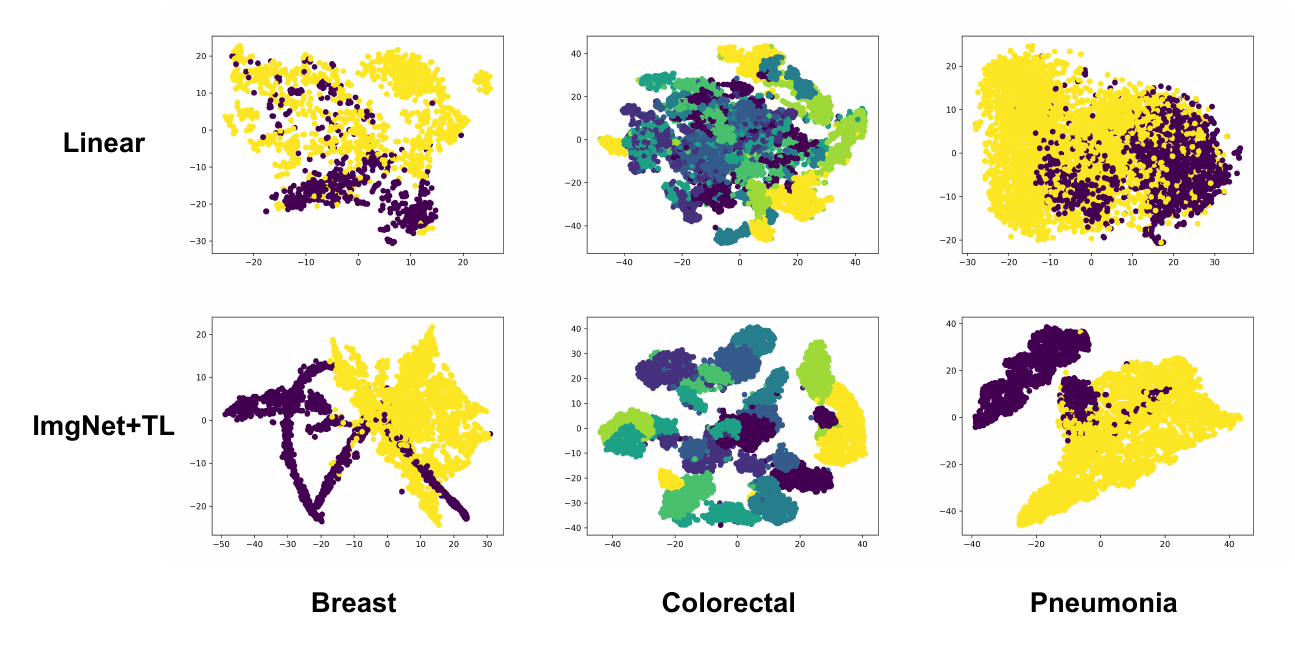}
    \caption{$t$-SNE comparison of top-performing SSL model vs supervised setting across all 3 datasets.}
    \label{fig:tsne}
\end{figure}

\keypoint{Effect of class imbalance, SSL vs Supervised:}SSL methods have been found to be \textit{intrinsically robust} to dataset imbalance \cite{liu2021self}. This is more appropriately shown by the consistently high $F1$-scores that keep parity with the accuracy values for imbalanced datasets (Breast and Pneumonia). In fact, we note that for Pneumonia, there is a large difference between $accuracy$ and $F1$-$score$ for the supervised setup $(\simeq 0.023)$ as compared to the top-peforming SSL algorithm $(\simeq 0.010)$. \autoref{tab:lin} also reveals that linear probing of self-supervised representations outperforms transfer learning of frozen supervised ImageNet \cite{deng2009imagenet} features. Qualitatively, the $t$-SNE plots in \autoref{fig:tsne} clearly depict the skewed distribution of the supervised model compared to a more uniform distribution obtained from SSL. Thus, it is logical to infer that class imbalance is handled more effectively by self-supervision, even under low-resource settings.

\section{CONCLUSION}
In this paper, for the first time, we explored the performance of SSL on small-scale medical imaging datasets. We holistically evaluated four such methods across three real-world biomedical datasets. From our findings, it seems fair to conclude that SSL methods, especially non-contrastive ones, can be effectively leveraged to low-resource medical imaging data. However, we also note that certain dataset-specific nuances alter the expected SSL trends, which requires further research. Tackling such biases can potentially lead to more robust and data-efficient SSL algorithms. 


\newpage

\bibliographystyle{IEEEbib}
\bibliography{strings,refs}

\end{document}